\documentclass[10pt, journal, a4paper]{IEEEtran}

\IEEEoverridecommandlockouts
\usepackage{balance}

\usepackage{cite}
\usepackage[switch]{lineno}
\usepackage{adjustbox}
\usepackage{algorithmic}
\usepackage{graphicx}
\usepackage{textcomp}
\usepackage{xcolor}

\usepackage[utf8]{inputenc}
\usepackage[T1]{fontenc}

\usepackage{color}

\definecolor{pblue}{rgb}{0.13,0.13,1}
\definecolor{pgreen}{rgb}{0,0.5,0}
\definecolor{pred}{rgb}{0.9,0,0}
\definecolor{pgrey}{rgb}{0.46,0.45,0.48}

\usepackage{listings}

\usepackage{hyperref}
\usepackage{booktabs}
\usepackage{color, colortbl}
\usepackage{subcaption}
\usepackage{fixltx2e}
\usepackage{url}
\usepackage{mathtools}
\usepackage[disable]{todonotes}
\usepackage{cleveref}
\usepackage{verbatim}
\usepackage{amsmath}
\usepackage{multirow}
\usepackage{lipsum}
\usepackage{mdframed}
\usepackage{textcomp}
\usepackage{float}

\definecolor{palegrey}{gray}{0.85}

\begin{document}
\title{Tailoring Requirements Engineering \\for Responsible AI}


\author{Walid Maalej, Yen Dieu Pham\thanks{W. Maalej and Y. Pham are with Universität Hamburg, Germany.} 
and Larissa Chazette
}

\maketitle

\begin{abstract}
Requirements Engineering (RE) is the discipline for identifying, analyzing, as well as ensuring the implementation and delivery of user, technical, and societal requirements. Recently reported issues concerning the acceptance of Artificial Intelligence (AI) solutions after deployment, e.g. in the medical, automotive, or scientific domains, stress the importance of RE for designing and delivering Responsible AI systems. In this paper, we argue that RE should not only be carefully conducted but also tailored for Responsible AI. We outline related challenges for research and practice.

\end{abstract}


\section*{Introduction}
\label{sec:intro}
A remarkably high number of AI solutions either do not make it to the production environment\footnote{Gartner estimates that by 2020 only 53\% of AI prototypes actually make it to production:  \url{https://www.gartner.com/en/newsroom/press-releases/2020-10-19-gartner-identifies-the-top-strategic-technology-trends-for-2021}} or fail after deployment. 
The reason is often the same: a missing or a bad understanding of user, technical, and societal requirements. For instance, Google \cite{Beede:CHI:2020} recently encountered major issues when deploying its large-scale, top-accuracy Machine Learning (ML) model for detecting diabetic retinopathy in Thai hospitals. 
Beede et al.~\cite{Beede:CHI:2020} observed that the detection accuracy decayed compared to the lab evaluations since $\sim20\%$ of the field data did not meet the image quality assumed during the  training and evaluation of the model. 
Moreover, there were serious user acceptance issues because the new system was poorly integrated into the existing hospital infrastructure, disrupted the nurses workflows, and compromised the patient experience causing a significant overhead. 
The authors argued: ``Currently, there are no requirements for AI systems to be evaluated through observational clinical studies, nor is it common practice. This is a problem because the success of a deep learning model does not rest solely on its accuracy, but also on its ability to improve patient care.'' 

In recent years, multiple similar examples of AI project failures resulted in negative \href{https://www.forbes.com/sites/larrymagid/2020/06/12/ibm-microsoft-and-amazon-not-letting-police-use-their-facial-recognition-technology/?sh=3b5566df1887}{news coverage} and brought serious consequences to the software vendors and to the society at large. 
Fry \cite{Fey:Book:2018} created a collection of prominent AI failures that raised serious societal or legal concerns, including medical ML systems that failed to provide the correct diagnosis with negative impact on people's health, \href{https://en.wikipedia.org/wiki/Death_of_Elaine_Herzberg}{accidents} involving self-driving cars that failed to recognize pedestrians and obstacles, or predictive policing systems that erroneously identified black people as being more likely to commit a crime \cite{Raji:AIES:2019}. 
What these cases have in common, is that their AI models are designed and trained in a lab environment, representing a limited understanding and representation of the real world scenarios, and not accurately reflecting the context when making a decision. 
Technology-driven AI solutions tend to prioritize automation over stakeholder needs and to oversimplify rare but important scenarios and tradeoffs. 
Moreover, a lack of transparency and explainability of AI-based solutions often lead to mistrust and low acceptance by users \cite{Chazette:RE:2021}.

\begin{figure*}[]
\centering
\includegraphics[scale=0.62]{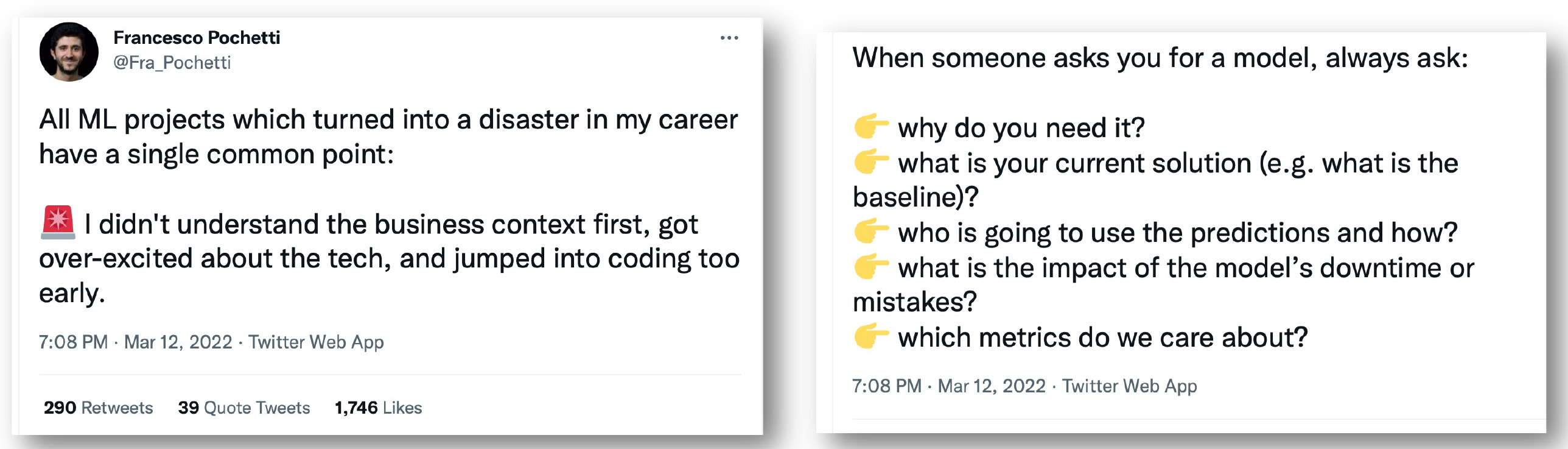}
\caption{A recent popular Tweet about the importance of Requirements Engineering for Machine Learning projects by Pochetti, who was named a “Machine Learning Hero” by \href{https://aws.amazon.com/developer/community/heroes/francesco-pochetti/?trk=dh_card}{AWS in 2019}.}
\label{fig:viraltweet}
\end{figure*}

While some might argue that these issues represent greater scientific, regulatory, and societal challenges, the good news for responsible AI is that there is already an established pragmatic engineering discipline which focuses on understanding stakeholder needs, specifying acceptable requirements, and ensuring the satisfaction of these requirements to a reasonable extent within the solution space. 
This discipline is called Requirements Engineering (RE) with a large body of knowledge that has emerged over the last 45 years of research and practice \cite{Gregory:Software:17} 
including a vivid research community, a fairly large practitioner community, several standards (such as IEEE 830), and certification bodies (such as \href{https://www.ireb.org/en/}{IREB} and \href{https://www.iiba.org}{IIBA}). There are multiple, well-studied, domain- and project-independent RE methods, templates, and tools \cite{Wiegers:Book:2014} that can also be used for Responsible AI Engineering projects, particularly to:  

\begin{itemize}
    \item Conduct interviews, workshops, or focus groups with stakeholders to a) identify the needs, requirements, boundary cases, and constraints concerning the system in general and the data in particular.
    \item Co-design incremental acceptable solutions with precise acceptance criteria.
    \item Run observational studies and as-is analysis, not only to understand the users’ workflows, technical and legal constraints, as well as the system interfaces, but also to specify the baselines against which new AI solutions should be compared.
    \item Evaluate the feasibility, priorities, and costs of requirements, while exploring and documenting the tradeoffs.
    \item Create user stories, use cases, empathy maps, (acceptance) test cases, or other  models in order to a) analyze, document, and validate the requirements for subsequent development and b) to check and trace the progress. 
\end{itemize}

\vspace{6pt}
However, there are also two bad news that challenge the application of RE in AI projects. 
First, RE is often a hard-to-budget, easy-to-ignore activity in software projects. While developers and stakeholders often recognize the importance of RE, the actual doing falls short. 
Dedicated roles of requirements engineers or business analysts as well as dedicated RE activities such as elicitation, analysis, traceability, or validation are often missing, unless they are required by contract or compliance rules, e.g., in governmental and aerospace projects. 
Therefore, one may argue that it is unrealistic to expect each AI project or AI organization to have dedicated RE professionals. In such cases, it is all the more important that, on the one hand, Responsible AI engineers acquire and understand basic RE knowledge to engage in RE activities by themselves; and on the other hand, that other roles (e.g., product owners, UX experts, scrum masters, quality and governance experts) collaborate with AI engineers from the early phases with specific emphasis on RE. For instance, based on a study of different organizations, Nahar et al.~\cite{Nahar:ICSE:22} suggested that the product team must involve the data science team in the negotiation of requirements to avoid unrealistic expectations.

Second, several recent studies of Software Engineering practice in ML projects \cite{Nahar:ICSE:22, Wan:TSE:19,Amershi:ICSE:19, Ishikawa:CESI:19} particularly highlighted the RE-related pains. In a survey with 278 practitioners, Ishikawa and Yoshioka \cite{Ishikawa:CESI:19} stated that most engineering difficulties encountered in ML projects were caused by insufficient understanding of customers or too much expectation. Wan et al.~\cite{Wan:TSE:19} interviewed 14 practitioners and found that most interviewees made strong statements about differences between the requirements of ML systems versus non-ML systems, yet these differences seem not fully understood by research. 
Interviewees stated that requirements are more uncertain for ML systems than non-ML systems, arguing that the goals are often too abstract (e.g. to automate or predict something). Unlike traditional systems which expose multiple logical states, ML systems expose a blackbox behavior that is hard to specify, analyze, and validate.

Motivated by these studies and to tap the full potential of RE in ensuring the deployment of Responsible AI systems that meet user, system, and societal requirements, we discuss six aspects that need particular attention and tailoring to the AI context. These aspects emerged from analyzing the specifics of the AI domain, the literature, and our own experience with AI projects (mainly ML and Data Science projects). 

\section{ACCEPTABLE LEVELS OF QUALITY REQUIREMENTS}
\label{sec:acceptable}
Quality requirements (sometimes also called non-functional requirements or system qualities) such as accuracy, explainability, scalability, adaptability, fairness, and error-tolerance  seem  particularly important and challenging for AI projects \cite{Horkoff:RE:19, Vogelsang:AIRE:19}. 
These qualities are generally fuzzy, hard to specify, and measure. Their importance depends not only on the domain (e.g., self-driving cars vs. weather prediction) but often also on the specific usage context and might even be subject to change depending on the stakeholders’ awareness \cite{Habibullah:RE:21}. 
Moreover, quality requirements are difficult to guarantee due to their dependency on multiple components, underlying technologies, external services, or possibly noisy field data. 
As a consequence, qualities might get only a low or a late attention in the projects, first when serious acceptance issues emerge or negative news coverage happens \cite{Raji:AIES:2019}. 

However, to ensure implementability, testability, and contractual compliance, qualities should be negotiated and specified in a precise and measurable way. For example, instead of the ambiguous requirement ``the system should be usable'', a certain time interval for performing a user task, a number of user interactions, or satisfaction scores for particular use cases should be specified. This can be very challenging in Responsible AI projects as quality goals tend to remain tacit. 
For instance, stakeholders often agree that a ML classifier should be as accurate as possible, after the motto: the more accurate the better. While this imprecise requirement might be desired for research papers, benchmarking studies, or open competitions, in the case of Responsible AI systems, accuracy requirements must be made as concrete as possible to stakeholders and AI engineers. 
That is, stakeholders should explore, negotiate, and specify for a certain use case what metrics for measuring a certain quality are important (e.g. precision, recall, or AU-ROC for classification accuracy) and what are the \textit{acceptable values}. This also guides the design and optimization of AI models.  

Due to stakeholder fuzzy expectations and technical uncertainties in early project phases, agreeing on precise values is difficult or unreasonable. 
Using \textit{Quality Levels} instead of fixed values can be a good compromise to overcome this challenge. 
For instance, AI engineers and stakeholders might agree on the following requirement: The acceptable precision for release 1 should be Level 8 (that is 80-89.9\%) and for release 2 Level 9 (>90\%) while the recall should be at Level 9 for release 1 and 10 for release 2 (>98\%). 
Standardizing quality levels similar to energy labels or food classifications can hide complex  technical specification details while still allowing for comparison and an approximation of what to expect. 
This increases the acceptance chances for AI systems as it helps clarify the user expectation and the comparability between solutions. 
What levels are reasonable for what qualities remains an open question for researchers and standardization bodies.

\begin{figure}[]

\centering
\includegraphics[scale=0.45]{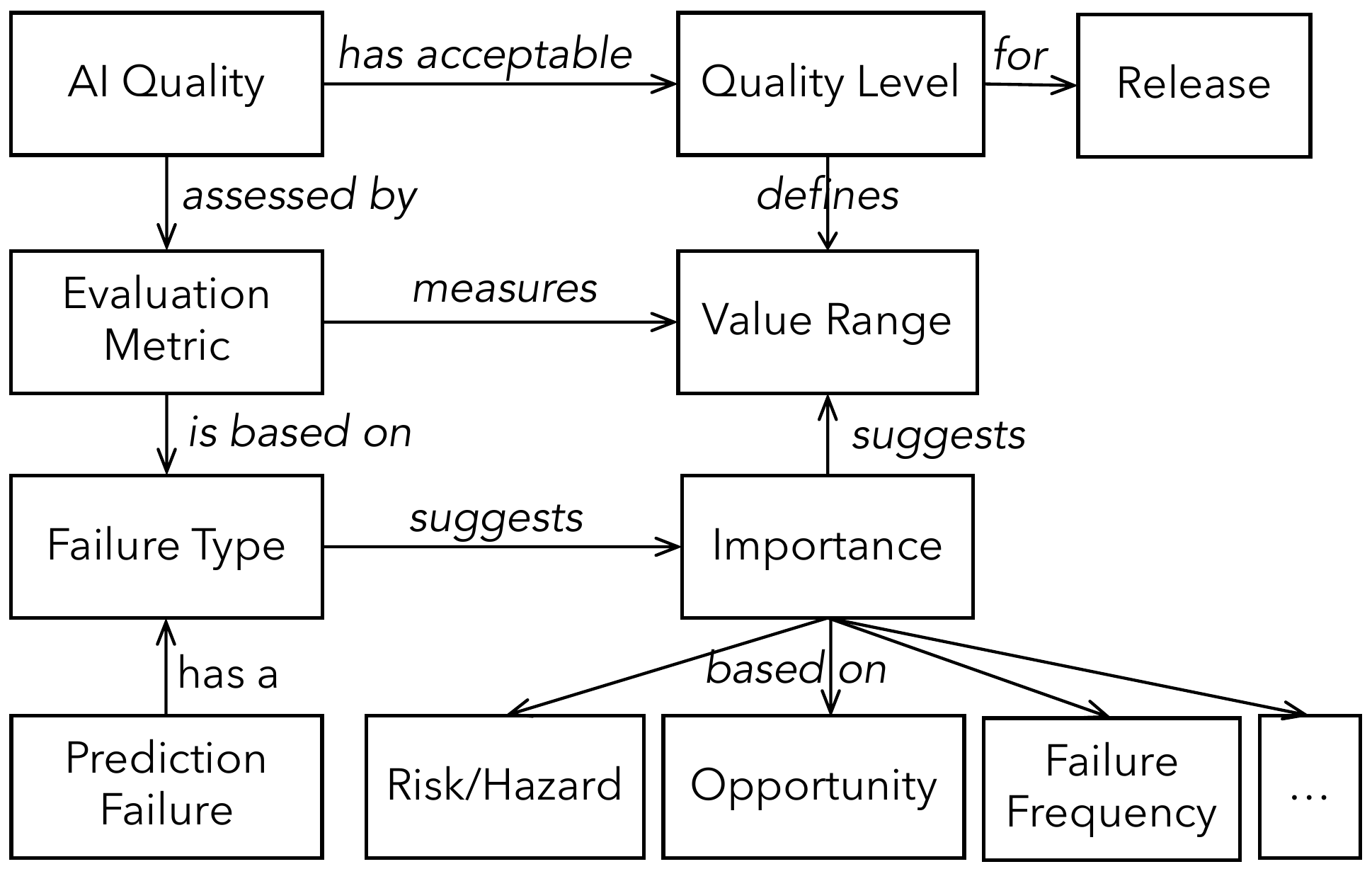}
\caption{A simple metamodel for discussing and specifying acceptable quality requirements.}
\label{fig:failure-metamodel}
\end{figure}

To identify acceptable levels of qualities, in addition to a domain and stakeholder analysis, carefully analyzing boundary conditions and exception handling of use cases as well as potential types of failures -- particularly \textit{prediction failures} -- is a key. Qualitative and quantitative systematic approaches for \textit{Failure Analysis}, including Failure Mode Analysis or Fault Tree Analysis (which are common in safety-critical domains such as aerospace, energy/nuclear plants, or medical devices) can serve as a starting point and should be evaluated and adjusted to the domain of responsible AI. 
Such analysis often include the identification of failures and failure types, the evaluation of their severity and probability, as well as the identification of possible hazards or impacts. 
This process can be prospective to guide responsible AI engineers   optimize their models based on what is more important and help  customers and other stakeholders understand the limitations and risks of the systems. 
Also fallback solutions, e.g., to include a human oracle in the decision loop can be designed. 
The baselines (i.e., what values are known from the state of the art or practice), development costs, speed, and other tradeoffs can play a role to define what is acceptable. 

Figure \ref{fig:failure-metamodel} depicts a simple metamodel that can structure AI failure analysis and the specification of acceptable quality levels. 
This analysis can particularly be guided by the following questions: 

\begin{itemize}
    \item What types of prediction failures can occur, when, at what frequency, and why? 
    \item What evaluation metrics are more important and should be optimized for and why?
    \item What is the cost or risk associated with the different types of prediction failures?
    \item When are particular failures (and resulting biases) acceptable to whom, at what average rate, at what frequencies, and in which legal, societal and user/task context? 
    \item Can quality metrics be decomposed into more specific use cases or prediction tasks (e.g., optimizing for the accuracy of certain (more important) classes instead of averages)? 
\end{itemize}
The answers to those questions can differ when comparing what stakeholders think and what they actually do and choose. Thus, interview studies and surveys (for capturing stakeholders’ opinions) as well as observational studies and A/B Testing (for capturing their behavior) might be needed.

\section{DATA- AND USER-CENTERED PROTOTYPING FOR AI}
\label{sec:prototyping}
Prototyping is traditionally an important technique not only to communicate with stakeholders, clarify their expectations, and gather their feedback, but also to explore the solution space and check the requirements feasibility. 
While prototyping is common in AI, it usually focuses on exploring the data, technology, and feasibility, e.g. through tuning and optimizing ML models. Data scientists and AI engineers often create Computational Notebooks to quickly explore datasets, train, and tune prediction models. 
A big advantage of Notebooks and data scripting environments in general is that they allow for quick exploration of the data and the prediction technology stack (i.e. data- and technology-centered prototyping) with workflow automation in mind (given a certain input, calculating a certain output).  
The disadvantage, however, is that they are typically decoupled from user-centered prototyping, which focuses on user tasks and interactions \cite{Lowdermilk:Book}.
Bridging these two prototyping perspectives, i.e. data- and user-centricity, is not trivial since the target groups and the tooling are different.

Prototypes that only focus on exploration and feasibility risk to remain with the AI teams and not get released to and evaluated with actual users. 
Such notebooks usually target research and scientific users showing summative or analytical visualizations of the data and the models rather than a system perspective in the user environment. Notebooks are also often self-contained and hard to integrate into other environments. 
Moreover, if the expectation is to fully automate certain decisions with the AI solution under development, demonstrating multiple erroneous predictions to users may lead to a strong expectation gap, leading in the worst case to losing the trust in the system and stop using it before it gets  released. 

Releasing AI models early to particular user groups, communicating transparently the level of qualities achieved, and offering a fallback solution (with a Human in the Loop to manually correct predictions, e.g., based on achieved prediction uncertainty \cite{Andersen:ACL:22})  increase acceptance and allow to test the AI models in the wild on unseen data. 
In the best case, this would enable AI engineers to collect additional data for model retraining and improvement. 
The challenge is to achieve a balance between the understanding of the problem space (including what to automate and how to integrate into the stakeholders workflows) and exploring the solution space (how and how well automation can be achieved).

\section{Expanding RE To FOCUS ON DATA}
\label{sec:redata}
Data plays a central role in AI, particularly in the subfields of ML and Data Science. Many properties and qualities of AI systems directly dependent on the properties and qualities of the data \cite{Strickland:Specctrum:22}. 
Traditionally, however, data plays a rather secondary role in RE. Instead, the system behavior as well as the system interaction with the user and the environment often represent the main focus of elicitation, analysis, and testing activities. 
Therefore, researchers and practitioners should consider adjusting RE activities, templates, and tools to focus more on data. This includes requirements for the collection, integration, pre-processing, labeling, enrichment, storage, usage, and sharing of data. 

Example questions for guiding Data RE include: 
\begin{itemize}
\item What data is available? What data is needed? 
\item What is acceptable for the stakeholders and the regulations? 
\item What technical constraints concerning the data compatibility and processing exist and how hard are they? 
\item Who should label the data and how? What labeling errors and labeling disagreements are acceptable in which scenarios? 
\item What consequences do labeling disagreements have for system requirements and qualities? 
\item What data sampling requirements are needed and what is feasible? 
\item What datasets can be used and what compliance and quality issues might emerge? 
\item How to assess the quality of the data and how to communicate it to the stakeholders (in the system)? 
\item What meta-data is needed and why? 
\item What data governance policy applies and who owns what data? 
\item What technical, legal, and ethical constraints are there (e.g. license, copyright, privacy requirements, consents etc.)?  
\end{itemize}

As for any project, the criticality of these questions, with whom they should be discussed, as well as other follow-up questions depend on the project goals, phase, as well as the level of knowledge available. These questions and resulting requirements are often interdependent and need to be identified, analyzed, linked, and tradeoff'ed (as discussed below). 
Current RE practices, such as the creation of use stories or use cases might need to be expanded to address Data RE. Shifting the focus to Data RE is an important research direction and is inline with the data-centric AI movement \cite{Strickland:Specctrum:22}, which is defined as “the discipline of systematically engineering the data needed to successfully build an AI system”.

\section{EMBEDDING RESPONSIBLE AI TERMINOLOGY INTO THE ENGINEERING WORKFLOWS}
\label{sec:embedding}
In their Ethics Guidelines for Trustworthy AI \cite{HLEG:Guidelines:19},
HLEG\footnote{\url{https://digital-strategy.ec.europa.eu/en/policies/expert-group-ai}}, an expert group appointed by the European Commission to provide advice on its AI strategy, suggests seven core requirements that are key to developing responsible, trustworthy AI. Those are: 1) human agency and oversight; 2) technical robustness and safety; 3) privacy and data governance; 4) transparency; 5) diversity, non-discrimination and fairness; 6) societal and environmental wellbeing; and 7) accountability.
The suggestions of the expert group was also based on feedback from the European AI Alliance, an online forum with over 4000 researchers, practitioners, and policymakers. 
The experts highlight the importance to operationalize and assess these core requirements in practice as they are high-level and hard to implement, trace, and audit. They also created a detailed guide for self-assessment of the requirements, \cite{HLEG:Assessment:20}, 
mainly consisting of a catalog of questions for each requirement. Yet the operationalization of these requirements in technical, implementable, and testable features is still an open challenge. 

In a recent study, Pham et al.~\cite{Pham:RE:21} studied the impact of \textit{exposing} stakeholders to a certain terminology on the resulting requirements. This phenomenon -- called the Linguistic Relativity Theory -- indicates that the vocabulary used by humans influences their thinking. 
In their study, the authors showed that embedding a sustainability vocabulary into requirements elicitation sessions leads to more concrete sustainability requirements compared to a control group not exposed to such vocabulary. 
We think that this simple idea is crucial for operationalizing Responsible AI in practice. 
That is, general system qualities like human agency and oversight, transparency, or sustainability, which engineers  find rather fuzzy and hard to operationalize in measurable way \cite{Pham:RE:21}, should be decomposed into dimensions (i.e.~the vocabulary) which should then be embedded into various RE activities, templates, and artifacts. 

For instance, sustainability might include environmental, social, and economic sustainability. Fairness might cover gender, profession, and ethnic fairness. 
Explainability might include visualization, communication, and interpretability aspects at different levels for different stakeholders. 
These dimensions constitute the simple vocabularies to which stakeholders and Responsible AI engineers should be exposed, e.g. in form of questions in elicitation interviews, discussion topics in design  workshops, sections in requirements specifications documents, or annotations for mockups and epics. This way, these goals become more accessible and concrete in early project phases to stakeholders and AI engineers who have to repeatedly think about them. The representation of the vocabulary (e.g., brief/long explanation, examples, icons, etc.) is also likely to make a difference.
In fact, the catalog of questions in the Assessment List for Trustworthy Artificial Intelligence (ALTAI) provided by HLEG can guide the systematic creation of such vocabularies for the seven core requirements.

\vspace{6pt}
Researchers should investigate how effective for operationalizing Responsible AI systems are certain terminologies and representations in certain project contexts with certain backgrounds of the AI team. For instance, researchers could investigate whether the word \textit{environmental} leads to more environment-related requirements or whether another word is more effective in the language space of AI engineers, such as \textit{energy efficiency}. For measuring the impact of the terminology and to facilitate auditing and documentation, the dimensions should be linked and traced back and forth to different project artifacts including source code and model documentation.  

\begin{figure*}[h!]
\centering
\includegraphics[scale=0.65]{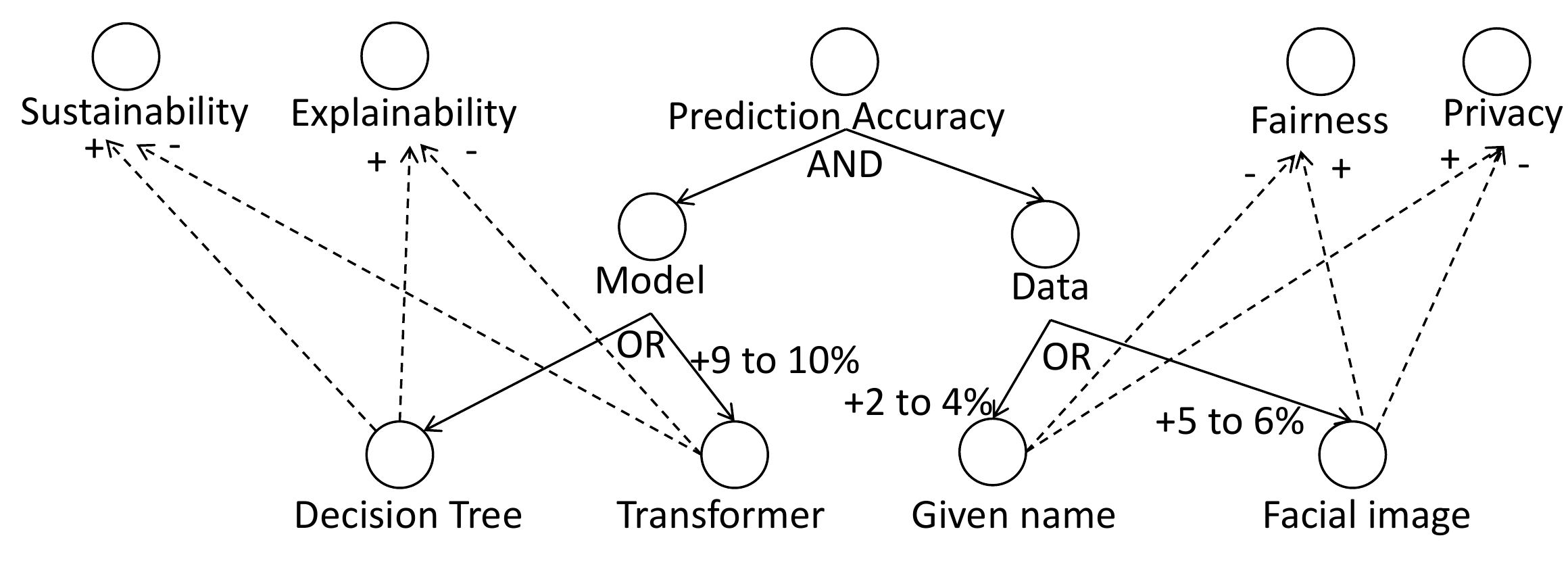}
\caption{An excerpt from a tradeoff model for an AI system that aims at recommending personalized cultural news.}
\label{fig:tradeoff}
\end{figure*}

\section{TRADEOFF ANALYSIS FOR RESPONSIBLE AI}
\label{sec:tradeoffs}

Tradeoff analysis is a central activity of requirements analysis \cite{Dutoit:Rationale:06, Chazette:ICSSP:22} 
since requirements, particularly quality requirements, are often in conflict. For example, a common tradeoff is between usability and security requirements \cite{Roh:REW:17}, as additionally required security measures (e.g., two-factor authentication) can lead to constraining the system usability (e.g., additional interaction steps). 
For Responsible AI systems, the prediction accuracy might be constrained with ecological sustainability requirements as analyzing large datasets, ML benchmarking, hyperparameter tuning, and retraining ML models often come with significant energy consumption and $CO_2$ emission \cite{Garcia-Martin:PDC:19}. A recent study suggests, for instance, that the $CO_2$ emission of building a Transformer-based language model for English with optimization and search can be compared to the total lifetime carbon footprint of five cars \cite{Strubell:ACL:19}. 

Other tradeoffs exist between explainability and usability when additional explanation dialogues are shown to the users, or between explainability and accuracy when, e.g., a neural network classifier achieves a higher precision than a decision tree classifier at the expense of tracing back and explaining the specific ML features that resulted in that particular prediction. 
Another common tradeoff example is between achieving fairness (or avoiding discrimination) and protecting the privacy of users, as additional datapoints about minority groups need to be collected to study and unbias a model. 
Tradeoff analysis helps Responsible AI engineers overview the decision landscape, prioritize, and document the decisions. 

\vspace{6pt}
There are multiple well-known tradeoff analysis techniques in RE, going from descriptive to prescriptive approaches, yet their suitability for the AI context is still to be shown. 
Figure \ref{fig:tradeoff} shows a simplified tradeoff analysis concerning a recommendation system for personalized cultural news. Some key quality requirements were identified (top left to right of the figure), including sustainability, explainability, prediction accuracy, fairness, and privacy. By making the tradeoffs explicit, practitioners can  prioritize requirements and design solutions. 
Design decisions with respect to the ML solution include the type of model and the data needed to train this model. 
These  decisions have different impacts on the other quality requirements. 
For instance, a Transformer model can lead to a better prediction accuracy, but has a worse explainability, costs more to retrain, and does not contribute to sustainability since it would generate a larger carbon footprint. The model shows common positive and negative assessments (shown as + and - in the figure for the corresponding qualities) as well as estimates the single design options would have on quality metrics (+/-x\%).

If the arguments for the pros and cons are borderline, not quantifiable, or purely subjective  reflecting stakeholders’ opinions and preferences, Responsible AI teams might consider negotiation and estimation techniques such as Planning Poker \cite{Alhamed:ICSE:21}. If the product has a large user base with a significant public interest, user involvement techniques such as panel surveys, feedback analytics, or open user discussion and voting (e.g. in online forums or user participation platforms such \href{https://uservoice.com/}{UserVoice}) can be useful not only for making a decision but also for reflecting and educating users about the difficulties of choices \cite{Johann:RE:15}. 

\vspace{6pt}
For Responsible AI, tradeoff analysis should be carefully conducted, as it is unlikely that all requirements can be achieved at once and at the same level of quality. Some requirements might also be subject to technology maturity or data availability. 
The European Ethics Guidelines for Trustworthy AI highlight the importance for analyzing tradeoffs and potentially including external stakeholders such as ethical boards. 
Tradeoff analysis in engineering projects, such as the example depicted in Figure \ref{fig:tradeoff}, helps to: 

\begin{itemize}
    \item Externalize tacit knowledge (i.e., why and how certain requirements are constrained, or what are the strengths and limitations of different options).
    \item  Create awareness and a common ground across the stakeholders (possibly noting unrealistic or unfeasible expectations).
    \item  Make and document rationale-based decisions (i.e., based on the assessment of concrete criteria and discussion of pros and cons).
\end{itemize}

It is also worth noting that tradeoff analysis can also be considered a design task as the solution space is explored and alternative technologies, models, or datasets are evaluated. It can also be considered a documentation and knowledge sharing task, as awareness of the tradeoffs, the constraints, strengths and limitations of different options are discussed and documented.

\section{REQUIREMENTS AS FOUNDATION FOR QUALITY AND TESTING Of AI}
\label{sec:testing}
The ultimate goal of RE, also for Responsible AI, is to ensure the continuous  delivery of system versions that satisfy users, supply their needs, comply with the regulations, and meet the expected quality.
It is therefore crucial that RE activities and artifacts impact and depend on other engineering and management activities -- particularly the continuous testing and quality assurance. 
Ideally, requirements should always lead and be linked to test cases.

Along with RE-related challenges, recent studies also highlighted major challenges with testing ML systems \cite{Nahar:ICSE:22, Wan:TSE:19,Amershi:ICSE:19, Ishikawa:CESI:19}. Because the system behavior is often non-deterministic, not all states of AI models can be specified, tested, and monitored as for other systems. 
What makes it even more challenging is that the term ``testing'' has its own meaning in the ML and Data Science communities, referring to a summative, simulation-based evaluation of the prediction accuracy of a trained model.
In addition to the \textit{model testing}, which is based on known data in the lab, Responsible AI engineers should carefully consider the black-box \textit{testing of the system and its parts}, which is conducted against specified requirements, often by different people than the developers, and following different techniques. 

Inspired by metamorphic testing, an established software engineering technique, Ribeiro et al.~\cite{Ribeiro:ACL:2020} recently introduced an approach called CheckList to test what they call ``individual capabilities'' of  NLP models. They developed a tool to help create and execute the test cases and evaluated it among others at Microsoft. 
The idea is to specify a list of qualities (or capabilities) expected from the model and to provide tool support for managing and testing these specifications. 
For instance, a sentiment analysis model should deliver the same sentiment, if in a sentence the name of a city, a person, or a country is changed (e.g. sentiment prediction for “Alice can’t lose her luggage moving to Brazil” should be the same as in “Bob can’t lose his luggage moving to Germany”) \cite{Ribeiro:ACL:2020}. 
The authors list many such capabilities for specific NLP models but do not explain where they originated from. 
We argue that these capabilities correspond to requirements either a) identified by human analysts, engineers, and other stakeholders; or b) reused from RE activities conducted in similar domains or  similar projects. 

Specifying and testing AI model capabilities should be an incremental process based on an initial analysis and on the feedback of stakeholders, e.g., in form or reviews and bug reports submitted after a release. 
Yet, research still needs to study and evaluate the effectiveness of specifying and reusing various capabilities in practice. 
Different types of systems might have different capabilities in different contexts. Moreover, capabilities will likely have tradeoffs as discussed in the previous section. For instance, requiring a certain level of 
 geographical fairness to the sentiment analysis of a text on loosing the luggage in the example above can be tradeoff'ed with a precision requirement to reflect a higher luggage lost frequency on certain routes or to reflect certain cultural tones.
Also the automation support for testing capabilities need further investigation, while ensuring that the human creativity (for discovering, specifying, and tweaking test cases) remains a central step within the loop. 

\section*{SUMMARY}
\label{sec:summary}
In summary, we think that the first step for operationalizing Responsible AI is to ensure that RE activities and techniques are known and thoroughly applied in AI projects. For taping the full potential, we have discussed six areas which need a particular attention by research and practice. Qualities of AI systems such as fairness, explainability, sustainability, or human agency are rather fuzzy and need to be analyzed with the various stakeholders. 
Systematic failure and tradeoff analysis can help set realistic expectations and specify acceptable yet feasible levels of qualities. Exposing the stakeholders and  engineers to common shared Responsible AI vocabularies  (e.g., in design meetings, prototypes specification documents, and feedback forms) helps create awareness and lead to precise AI models capabilities and system requirements. These should then be translated into test cases and continuously checked and expanded to  ensure a long term acceptance of AI systems outside the lab. 

\vspace{10pt}
As these areas are mainly based on the authors personal observations and experience, we refrain from claiming completeness and generalizability. 
There might be indeed other important challenging RE-specific aspects which can emerge, e.g. from interview or observational studies in specific domains. 
Also the outlined direction need to be investigated in detail, their applicability proved, and evaluated, e.g., in case studies or user studies. 
Therefore, collecting and comparing empirical studies and experience reports is perhaps a seventh focus area and challenge to the community at large for maximizing the benefits of RE for Responsible AI.



\IEEEtriggeratref{21}




\section*{ABOUT THE AUTHORS}
\vspace{6pt}
\noindent\textbf{Walid MAALEJ} is a professor of informatics at Universität Hamburg with focus on Empirical Software Engineering and AI Engineering. He currently serves as Steering Committee Chair of the Requirements Engineering Conference Series. Contact him at walid.maalej@uni-hamburg.de 

\vspace{6pt}
\noindent\textbf{Yen Dieu PHAM} is a certified building architect. She recently defended her PhD on Requirements Engineering and Sustainable Software Design at Universität Hamburg. Contact her at pham@informatik.uni-hamburg.de 

\vspace{6pt}
\noindent\textbf{Larissa CHAZETTE} recently defended her doctoral thesis on Requirements Engineering and Explainability at the Leibniz University Hannover, Germany. Contact her at lc@ntms.eu  

\end{document}